# Detection of Alzheimer's Disease Using Graph-Regularized Convolutional Neural Network Based on Structural Similarity Learning of Brain Magnetic Resonance Images

Kuo Yang, Emad A. Mohammed, Behrouz H. Far

*Abstract— Objective:* This paper presents an Alzheimer's disease (AD) detection method based on learning structural similarity between Magnetic Resonance Images (MRIs) and representing this similarity as a graph. *Methods:* We construct the similarity graph using embedded features of the input image (i.e., Non-Demented (ND), Very Mild Demented (VMD), Mild Demented (MD), and Moderated Demented (MDTD)). We experiment and compare different dimension-reduction and clustering algorithms to construct the best similarity graph to capture the similarity between the same class images using the cosine distance as a similarity measure. We utilize the similarity graph to present (sample) the training data to a convolutional neural network (CNN). We use the similarity graph as a regularizer in the loss function of a CNN model to minimize the distance between the input images and their k-nearest neighbours in the similarity graph while minimizing the categorical cross-entropy loss between the training image predictions and the actual image class labels. *Results:* We conduct extensive experiments with several pre-trained CNN models and compare the results to other methods. *Conclusion:* Our method achieves superior performance on the testing dataset (accuracy = 0.986, area under receiver operating characteristics curve = 0.998, F1 measure = 0.987). *Significance:* The classification results show an improvement in the prediction accuracy compared to the other methods. We release all the code used in our experiments to encourage reproducible research in this area [1].

## I. INTRODUCTION

Alzheimer's disease (AD) is the sixth cause of death in the United States [29]. AD is also the fifth cause of death for females and the ninth for males in Canada[2]. AD is a complex mental disease damaging patients' memory, communication, and thinking ability (i.e., dementia). Dementia has a wide range of severity from the mildest stage (i.e., affect a person's functioning) to the most severe stage (i.e., loss of ability to perform necessary daily life functions such as reading, spelling, and urination control). The causes of dementia can vary, depending on the brain changes [1]. The latest survey shows that AD has been a common cause for 60%-80% of dementia, affecting millions of people aged 65 or older [1]. However, there is no known cure for AD, and all current treatments are to slow down the AD-related symptoms.

Early detection of AD using deep learning techniques offers several medical and financial benefits. Deep learning has attracted increasing attention due to its success in image classification and Natural Language Processing (NLP). Magnetic Resonance Imaging (MRI) devices provide a massive amount of medical image datasets to harness the power of deep learning methodologies such as Convolutional Neural Network (CNN) and Recurrent Neural Network (RNN) in the early-stage detection of AD. Recent research on AD detection using deep learning methodologies mainly focused on fine-tuning the model's architecture [2-7]. Moreover, MRI image segmentation and classification CNN models were implemented to study specific brain regions related to AD [8,6]. Large high-quality labelled MRI image datasets constitute a significant challenge to achieve accurate AD predictions using CNN models. Thus, the CNN models were not able to learn relationships among individual MRIs and were not generalizable. None of the above researches studied the effect of the data balancing techniques on the model performance as an imbalanced dataset could significantly bias the model prediction.

Ensemble Learning methods were applied for the AD early-stage prediction to boost the model performance [9-12]. Some researchers contributed to the analysis of AD progression using RNN to model AD's monthly progression into the future, which is a multi-step prediction problem [13].

In this paper, we propose a graph-based method that enables CNN to learn both MRIs and their neighbouring relationships for the AD prediction, inspired by Neural Graph Learning (NGL) [14] and label propagation [15,16]. Our method starts with a data pre-processing and structural similarity learning pipeline to construct a similarity graph, map the graph nodes to the original MRIs, and end with the CNN model prediction.

Our method offers significant improvements and contributions to apply a graph-regularized method to the AD prediction. The main contributions of this paper are:

K. Yang is with the School of Engineering, Lakehead University, Thunder Bay, ON, P7B5E1, Canada (e-mail: kyang3@lakeheadu.ca). Emad A. Mohammed is with the School of Engineering, Lakehead University, Thunder Bay, ON, P7B5E1, Canada (e-mail: emohamme@lakeheadu.ca). Behrouz H. Far is with the Schulich School of Engineering, University of Calgary, Calgary, AB T2N 1N4, Canada (e-mail: far@ucalgary.ca)

1. We propose constructing and utilizing a similarity graph as an unsupervised learning method with a supervised learning method to predict AD using various pre-trained CNN models.
2. We use the graph to present (sample) the training data to the CNN models during training. The graph is used to randomly sample an image/embedding (seed) from the graph and then augment the nearest $k$ − neighbours with the seed. The neighbours are selected based on the cosine similarity. The sampling process is repeated until all training images are presented (i.e., used in the CNN training process) to the CNN models as seeds and k-neighbours. The similarity graph regularizes the CNN model weights by preserving each class images' structural similarity. The random presentation of the training images (order of the training images/seeds) maximizes the CNN model's ability to learn the classes' variations. Thus, the trained CNN model can be generalizable to various testing datasets within the class feature distribution. The CNN model loss (objective) function is composed of two components. The first component is the categorical cross-entropy that minimizes the average loss between the actual labels and the training images' predictions. The second component is the average cosine distance between the training image features and their k-nearest neighbour features.
3. We experiment with different methods to construct the similarity graph and test each method's effect on the models' performance.
4. We perform extensive experiments and compare our method for AD prediction with several current methods. The experimental results show that our method outperforms these methods.

This paper is organized as follows. Section II introduces the related work to our study; section III describes the MRI dataset, dataset balancing technique, and the similarity graph generation methods. Moreover, in Section III, our method is described in detail, including the background information, CNN model architectures, and objective function. Section V discusses the conducted experiments' results and the models' performance. We also further compare our study to recent work in terms of prediction accuracy and other quality matrices. Finally, Section VI is there to conclude our paper and discuss the limitation and future works.

## II. Literature Review

There are limited detection and diagnosis options for AD. They are based on either the General Practitioner Cognitive Assessment (GPCOG) [30], identifying AD biomarker from the Cerebro-Spinal Fluid (CSF) [31], or diagnosing MRI images. The literature review section focuses on the most recent studies using deep learning methods to detect AD from MRI image diagnosis.

A method based on an enhanced version of the CNN model known as the Inception V3 model was proposed to analyze MRI images to identify AD [2]. This method utilized three enhanced Inception blocks to improve the recognition accuracy of the Inception V3 model. The enhanced Inception V3 model achieved an accuracy of 85.7% in classifying patients with normal MRI, mild cognitive impairment (MCI), and AD using a total of 662 three-dimensional brain MRI images for training and testing.

Multiple cluster dense CNNs (DenseNets) method was proposed to classify MRI images for AD diagnosis [9]. The MRI images were portioned and cluster into different groups using the K-Means clustering algorithm. The DenseNets were then used to learn and classify features for each cluster. This method was evaluated using the Alzheimer's Disease Neuroimaging Initiative (ADNI) database[3] and achieved an accuracy of 89.5% for AD vs. Normal Control (NC) classification and 73.8% for MCI vs. NC classification.

A method based on an ensemble of CNNs was proposed to classify MRI images to identify patients with AD, MCI, and NC [10]. This method utilized a mixture of CNN architectures, including DenseNet121, Densenet169, DenseNet201, and ResNet50. This ensemble was evaluated using the structural MRI images from the Open Access Series of Imaging Studies (OASIS)[4] database. The results showed that the ensemble model achieved an accuracy of 95.23%.

A combination of U-net-like CNNs and the multinomial logistic regression method was proposed to identify and classify 3D MRI images' attention features for AD diagnosis [11]. The CNNs were used to learn the intra-slice features of the 3D MRI images, and multinomial logistic regression was employed to classify these features and identify patients with AD. This method was evaluated using the ADNI database and produced an accuracy of 97%.

An ensemble of CNNs was proposed to combine multiple MRI projections with different CNN architectures for AD diagnosis with high accuracy by learning the optimal fusion weights [12]. The authors designed an ensemble loss function to consider the interaction between the individual CNN models during the optimal weight search. This method was evaluated using the ADNI database and produced an accuracy of 94%.

A transfer learning-based method based on the VGG net was proposed for AD from MRI images [27]. The authors demonstrated that embracing a robust architecture for image classification through transfer learning and intelligent training data selection can improve AD identification accuracy. This method was validated on the ADNI dataset of 50 subject scans from each AD, MCI, and NC category and achieved 95% accuracy.

A multi-modality data fusion and classification method based on an ensemble of CNNs was proposed for AD diagnosis [28]. The authors employed the dropout mechanism and Adaboost algorithm to boost the accuracy of the ensembled model. This method was compared to the conventional machine learning algorithms and validated by comparing both the single modality and multi-modality data and achieved an accuracy of 91% in classifying AD vs. NC vs. MCI.

A transfer learning-based method was proposed to detect AD from structural MRI images [25]. This method utilized

---

[3] http://adni.loni.usc.edu/
[4] https://www.oasis-brains.org/

VGG16 and Inception V4 architectures with a small number of training images to obtain highly accurate AD predictions. The authors used an entropy-based technique to select the most informative images in a small MRI image set. This method was validated on images from the OASIS brain imaging dataset and achieved an accuracy of 96.25%.

A method for automated AD detection was proposed based on the Inception V4 CNN model for automated AD detection and classification [32,33]. The authors redesigned the final SoftMax layer for four different output classes (ND, VM, mild and moderate AD) and modified the Inception B and C modules so that they can accept the MRI data. The authors trained and tested their model using the OASIS dataset and achieved 73.75%.

Detection of AD is challenging due to the similarity of MRI images between AD and healthy patients. Several studies explored the AD diagnosis from MRI images. However, they focused on modifying and enhancing several CNN architectures or ensembled CNN models to produce high accuracy predictions for AD diagnosis. Our method focuses on highlighting the structural similarity of AD image classes (i.e., Non-Demented (ND), Very Mild Demented (VMD), Mild Demented (MD), and Moderated Demented (MDTD)) while maximizing the variance between classes to achieve robust and accurate predictions for AD diagnosis.

## III. MATERIAL AND METHOD

Fig. 1 shows the analytical pipeline used to pre-process and predict AD from MRI images. We first pre-process the images to balance the different classes and reduce the input image's features dimensionality, and generate a low dimensional embedding vector for each image. We cluster the low dimensional embeddings using the K-Means and hierarchical clustering algorithms based on the cosine similarity measure. We further construct a structural similarity graph representing each class of the input image separately and then merging them to build a structural similarity graph for all classes. The structural similarity graph serves two purposes: 1) we use the structural similarity graph to sample the training images and their embeddings by randomly selecting a training image (with its embedding) followed by its k-nearest neighbours from the graph and repeat the sampling process till selecting all training images. 2) we use the structural similarity graph as a regularizer term in the CNN model's loss function to keep the distance between the training sample embeddings and their k-nearest neighbour (i.e., minimizing the neighbour loss).

### A. Alzheimer's Disease MRI Images Data

The structural MRI data used in our paper is acquired from the OASIS database. The original dataset is composed of three sub-datasets (i.e., OASIS-1, OASIS-2, and OASIS-3). Fig. 2 shows samples of MRI images of (i.e. ND, VMD, MD, and MDTD). The OASIS-3 dataset provides the cross-sectional brain scans from ND, VMD, MD, and MDTD of different patients. The most challenging aspect of this dataset is that it has an imbalanced amount of MRI images between ND and MDTD, as shown in Table I. To generate high accuracy predictions of AD, we balanced the OASIS-3 MRI dataset using a combination of bootstrap and image augmentation methods to up-sample (increase) the minority classes and decrease model overfitting [23]. However, the up-sampling of the minority classes changes the prevalence of these classes and may drastically increase or decrease the similarity between a given MRI image (seed) and its neighbours.

TABLE I
MRI COUNTS OF EACH CLASS BEFORE AND AFTER BALANCING

| Class | ND | VMD | MD | MDTD |
|---|---|---|---|---|
| **Before Balance** | 3210 | 2240 | 896 | 64 |
| **After Balance** | 3210 | 3210 | 3210 | 3210 |

Therefore, our method's pre-processing part adds a small amount of standard Gaussian noise to the MRI images, followed by image saturation adjustment, left and right flip, slight rotation, and center crop. Finally, the newly generated images are shuffled back to the original dataset. The MRI image augmentation steps are shown in Fig. 3. The imbalanced data bias the model prediction to the majority classes such that the minority classes are not well modelled into the final model [24]. The data augmentation and bootstrapping balancing method can solve the imbalanced classes issue and add a particular variety to the minority classes. Image entropy analysis was utilized to filter the MRI images with high information [25]. This method is beneficial for the slightly imbalanced dataset but not for the heavily imbalanced MRI images dataset in our study. Selected samples from the balanced dataset are shown in Fig. 4.

### B. Dimensionality Reduction

#### 1) Student t-distribution Stochastic Neighbor Embedding

The Student t-distribution Stochastic Neighbor Embedding (t-SNE) was proposed to visualize the high-dimensional data by reducing the original data to 2 or 3 dimensions [17]. It was initially developed based on the Stochastic Neighbor Embedding (SNE). SNE used a transformation of the Euclidean distance of high-dimensional data into conditional probabilities [18] as follows:

$$p_{j|i} = \frac{\exp\left(-||x_i - x_y||^2 / 2\sigma_i^2\right)}{\sum_{i \neq k} \exp\left(-||x_i - x_k||^2 / 2\sigma_i^2\right)} \tag{1}$$

$$q_{j|i} = \frac{\exp\left(-||y_i - y_j||^2\right)}{\sum_{i \neq k} \exp\left(-||y_i - y_k||^2\right)} \tag{2}$$

The probability $p_{j|i}$ refers to the conditional similarity between the data point $x_j$ in the low dimensional space and the given point $x_i$ in the original data (high dimensional space) and the corresponding probability $q_{j|i}$ in the mapped points $y_i$. The neighbour points $x_j$ are collected under the Gaussian distribution with the mean $x_i$ and variance $\sigma_i$. The points $y_i$ are low-dimension weights used to match the distribution of high-dimensional data points $x_i$. Therefore, the loss function of the SNE algorithm is to minimize the sum of Kullback-Leibler divergences of the data points' conditional probabilities and map points through the gradient descent on the low-dimensional map points, as shown in (3).

$$Loss = \underset{y_i}{arg\min} \sum_i KL(p_{j|i}||q_{j|i}) =$$

$$\underset{y_i}{arg\min} \sum_i \sum_j p_{j|i} \log\left(\frac{p_{j|i}}{q_{j|i}}\right) \tag{3}$$

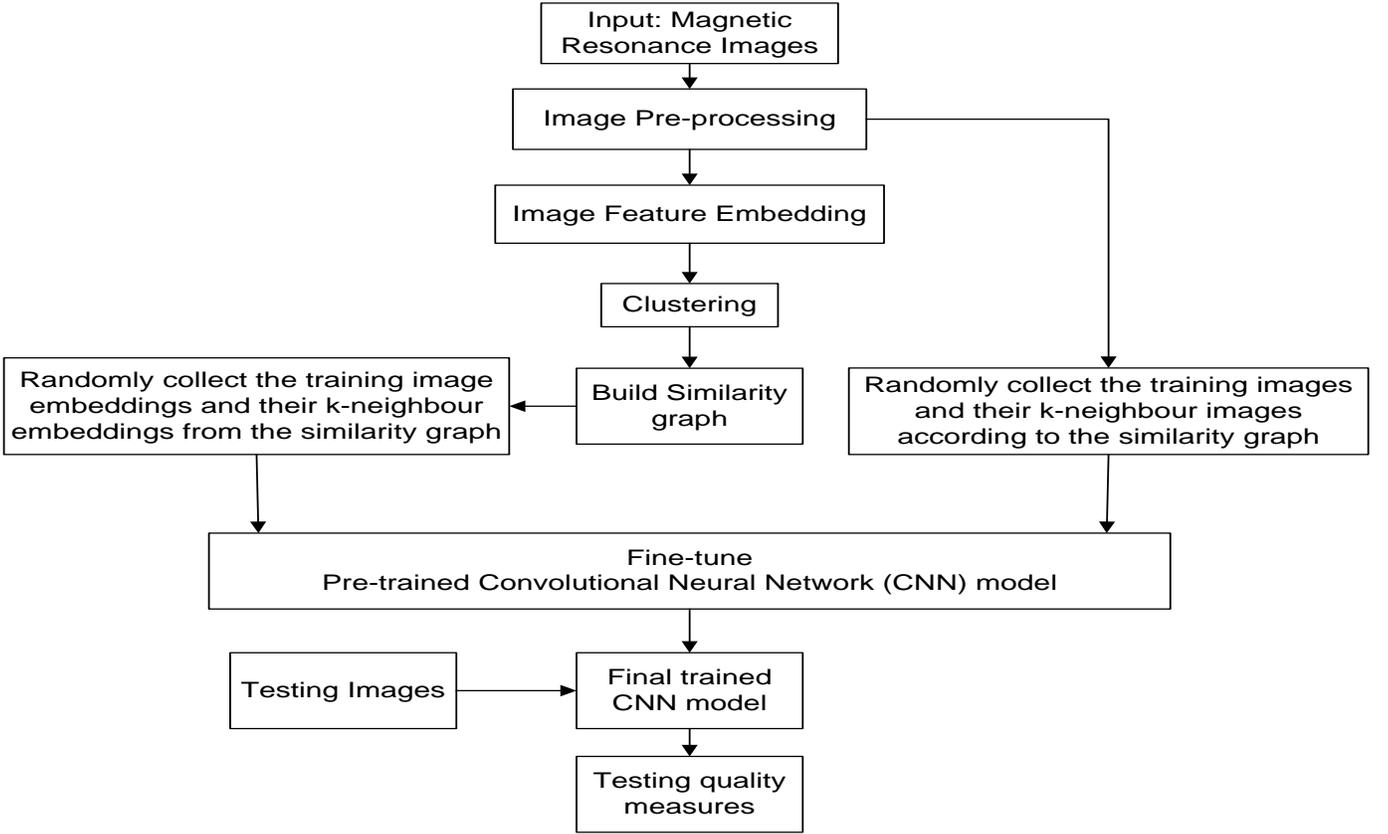

Fig. 1. Analytical pipeline for detecting Alzheimer's disease from brain magnetic resonance images using graph-regularized convolutional neural network based on structural similarity learning.

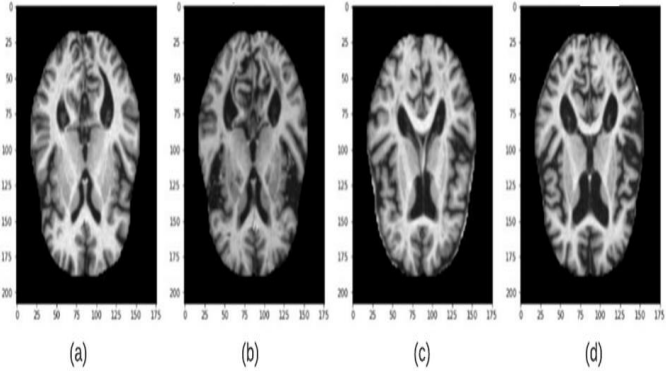

Fig. 2. Sample MRIs from OASIS-3 Dataset. (a) None Demented (ND), (b) Very Mild Demented (VMD), (b) Mild Demented (MD), and (d) Moderated Demented (MDTD))

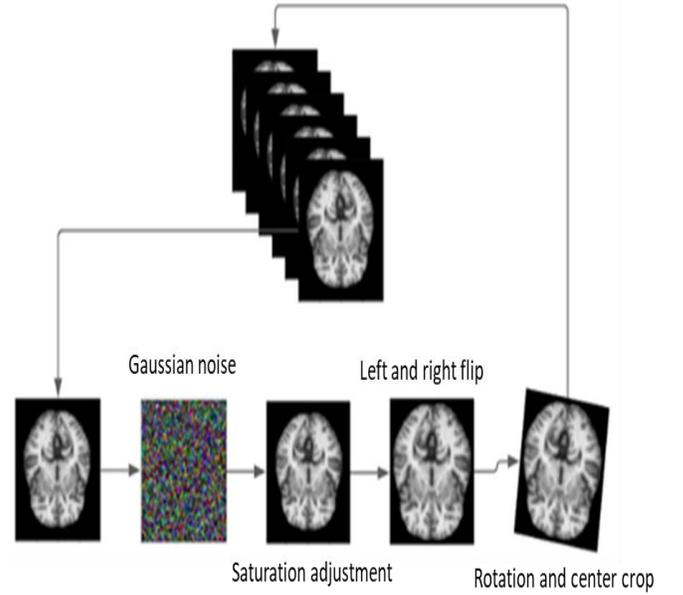

Fig. 3. MRI Augmentation steps

Compared to SNE, t-SNE keeps the same objective function, but it updates the conditional probability $p_{j|i}$, $q_{j|i}$ to the joint probability $p_{ji}$, $q_{ji}$. Instead of Gaussian distribution, t-SNE applies Student-t distribution with heavier tails to compute the similarity in low-dimensional map points. However, the similarity of high-dimensional data points keeps the same calculation as SNE. Therefore, the similarity and loss functions are updated as follows:

$$p_{ji} = \frac{\exp\left(-||x_i - x_y||^2/2\sigma^2\right)}{\sum_{k \neq l} \exp\left(-||x_k - x_l||^2/2\sigma^2\right)} \quad (4)$$

$$q_{ji} = \frac{\left(-||y_i - y_j||^2\right)}{\sum_{k \neq l} \exp\left(-||y_k - y_l||^2\right)} \quad (5)$$

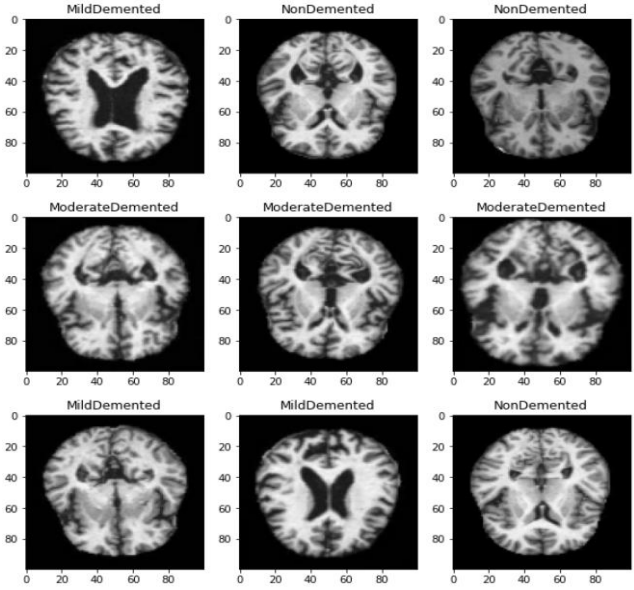

Fig. 4. MRI samples from the balanced dataset

$$Loss = arg \min_{y_i} \sum_i KL(p_{ji}||q_{ji}) = arg \min_{y_i} \sum_i \sum_j p_{ji} \log\left(\frac{p_{ji}}{q_{ji}}\right) \quad (6)$$

In our study, we use the t-SNE method to reduce the dimension of MRI images. If the mapping between low and high-dimensional data is correct, then the conditional or joint probabilities should be very closed [17]. Hence the map points $y_i$ can maintain the structure of the original data points. In our method we experiment with the t-SNE as dimensionality reduction and embedding technique.

### 2) Variational Autoencoder

Variational Autoencoder (VAE) [19] is a type of deep generative models for probabilistic machine learning. VAE is based on the Autoencoder (AE) algorithm [20] and is composed of an encoder and decoder. The encoder compresses the input data to low-dimensional representations, which can regenerate a similar input using the decoder. The VAE objective function is to minimize the difference between the original input and the reconstructed input. The VAE algorithm focuses on learning latent structures of the input; however, the AE algorithm ignores the input's probability distribution.

In contrast, the VAE algorithm constructs a latent space and then samples representations from this space to reconstruct the input. The VAE algorithm learns the statistical parameters of the latent space. In our study, we consider $x$ as the input from original input space, $z$ as the latent vector which is generated with the given input $x$ into the probabilistic encoder with weights $\varphi$ and $P_\varphi(z|x)$ as the distribution of latent space through the encoder in the VAE algorithm. The distribution of the input $P_X(x)$ is usually unknown or difficult to solve, and this makes the distribution of latent space $P_\varphi(z|x)$ hard to achieve either. The VAE algorithm approximates $P_\varphi(z|x)$ by $Q_\varphi(z|x)$ which is a Gaussian distribution. From this perspective, the VAE algorithm converts the calculation of the latent space distribution to an optimization problem, which can be solved by minimizing the difference between $P_\varphi(z|x)$ and $Q_\varphi(z|x)$, and also $Q_\varphi(z|x)$ can be simplified as a standard Normal distribution $N(0,1)$. Therefore, the final form of VAE's objective function is shown as $L_I$ in (7).

$$L_I = \underset{z \sim Q_\varphi(z|x)}{-E}[P_\theta(x|z)] + D_{KL}(Q_\varphi(z|x)||P_\theta(z)) \quad (7)$$

where $P_\theta(x|z)$ refers to the distribution of reconstructed input given the latent vector $z$ using a probabilistic decoder with weights as $\theta$. $P_\theta(z)$ refers to the distribution of the latent vector $z$ applied to the decoder. Equation (7) is the final form of the loss function. We add a reconstruction loss function to improve the reconstructed MRI images' quality, as shown in (8).

$$L_{II} = \frac{1}{n}\sum_{i=1}^n [x^{(i)} - f_\varphi(g_\theta(x^{(i)}))]^2 \quad (8)$$

where $f_\varphi$ refers to the encoder and $g_\theta$ refers to the decoder. Equation (8) is to compute the difference between the input and reconstrued input using the Euclidean distance. The final total loss function is illustrated in (9), and the total loss is used to update all trainable weights of VAE using the backpropagation method.

$$L_{total} = L_I + L_{II} \quad (9)$$

The main advantage of VAE is reducing the input dimension and maintaining the probabilistic features of input in the latent space. Therefore, we compress the MRI images from $100 \times 100$ (10,000) to a low-dimension space of 128 using the VAE algorithm. We choose a low-dimension space of 128 to suit the memory limitation of the cloud computing environment we use for computations (Google Colab[5]).

### 3) Adversarial Autoencoder

Like VAE, Adversarial Autoencoder (AAE) is also a probabilistic model with an encoder and decoder. However, the architecture of AAE is combined with the Generative Adversarial Networks (GAN) to perform variational inference [21]. GAN is a framework composed of a generator and discriminator models, and both can be constructed through the Multilayer Perception (MLP) [22]. The generator model indirectly learns the distribution of the real input data to generate fake inputs. The discriminator directly learns from real training data to predict the probability of input to be a fake or real input, and the GAN framework applies a max-min loss function to realize the competing process through the backpropagation during training [22] as shown in (10).

$$Loss = \min_G \max_D \{\underset{x \sim P_{data}(x)}{E}[\log(D(x))] + \underset{z \sim P_Z(z)}{E}[1 - \log(D(G(z)))]\} \quad (10)$$

where $P_{data}$ refers to the training data distribution and $P_Z$ refers to the sample data distribution. $D(x)$ refers to the discriminative model and $G(z)$ refers to the generative model. Equation (10) may cause a vanishing gradient issue since the second part of the (10) "E$[1 - \log(D(G(z)))]$" has shallow gradients when minimizing $D(G(z))$. GAN may be stuck at the early-stage of the training due to the discriminator's easy job [22], and (10) is updated in (11).

---

[5] https://colab.research.google.com/notebooks/intro.ipynb

$$Loss = \max_{G} \max_{D} \{ \underset{x \sim P_{data}(x)}{E} [\log(D(x))] + \underset{z \sim P_Z(z)}{E} [\log(D(G(z)))] \} \quad (11)$$

Therefore, AAE is constructed by combining both VAE and GAN into the network architecture. AAE constructs the latent space z with the distribution $Q_\varphi(z|x)$ by the reparametrization technique such as VAE in the encoder model $f_\varphi$ with the original input $x$. The generator model $g_\theta$ reconstructs the input $f_\varphi(g_\theta(x))$ from latent space samples. However, AAE maps the latent space distribution $Q_\varphi(z|x)$ to any arbitrary distribution $Q(z')$ using GAN. The generative model $G(z')$ takes samples from $Q(z')$ to confuse the discriminator, which the discriminative model $D(z)$ is learning from $Q_\varphi(z|x)$ to make a judgement. Thus, the posterior distribution $Q_\varphi(z|x)$ is no longer constrained to be a Gaussian distribution, and the encoder can learn any posterior distribution for a given input $x$ [21]. Our method applies AAE to reduce the MRI image dimension, and the loss function used in our experiment includes (8) and (11) in the final loss function of the CNN model under test.

### 4) Neural Graph Learning

Our method is primarily inspired by Neural Graph Learning (NGL) and graph-based semi-supervised learning algorithms. NGL involves a training framework that combines the power of label propagation and neural networks [14]. Label propagation is a semi-supervised and graph-based algorithm that constructs connections over the labelled and unlabeled data nodes in the graph. The known labels can then inference the information through the graph to label all the unlabeled nodes [15].

The graph-based semi-supervised learning produces a soft assignment of labels to each node in the graph [16]. Given a graph $G = (V, E, W)$ with $V$ nodes, $E$ edges and $W$ weights on the nodes, the convex objective function of the graph is shown in (12).

$$L = \mu_1 \sum_{v \in V_l} S_{vv} ||\hat{Y}_v - Y_v||_2^2 + \mu_2 \sum_{v \in V, u \in N(v)} w_{uv} ||\hat{Y}_v - Y_u||_2^2 + \mu_3 \sum_{v \in V} ||\hat{Y}_v - U||_2^2 \quad (12)$$

Equation (12) is subjected to $\sum_{l=1}^{L} \hat{Y}_{ul} = 1$, where $L$ is the total size of the node labels, and $\hat{Y}_{ul}$ is the soft assignment of the node labels. $S_{vv}$ is a seed node matrix and $S_{vv} = 1$ when node $v$ is the seed node. The $\hat{Y}_v$ is the soft assignment of labels using the training label $Y_v$ of seed nodes. $u$ refers to nodes in the neighbour of $v$ as $N(v)$. There are three hyperparameters $\mu_1, \mu_2, \mu_3$ to control the three measures in the objective function, including the distance between propagated labels and training labels, the distance between propagated labels and the neighbour labels under the same seed nodes with the edge weight, and the distance between the propagated labels, and the prior label distribution.

The seed label is predicted using MLP or CNN by minimizing the network supervised (i.e., label) loss function. Instead of assigning soft labels through propagation, NGL applies the same neural network to learn the seed node and neighbours' hidden representations at any intermediate or final prediction layers.

The distance between the representations is measured with the edge weight as the second term in (12) (i.e., unsupervised loss function). Finally, both network label loss and the node label distance loss (supervised and unsupervised) are summed up and applied to update the model weights using the backpropagation algorithm. However, to leverage the power of semi-supervised learning, the NGL objective function includes three different edge states between seed nodes and their neighbours, as shown in (13).

$$L_{NGL} = \mu_1 \sum_{(u,v) \in e_{ll}} w_{uv} d(h_\theta(x_u), h_\theta(x_v)) + \mu_2 \sum_{(u,v) \in e_{lu}} w_{uv} d(h_\theta(x_u), h_\theta(x_v)) + \mu_3 \sum_{(u,v) \in e_{uu}} w_{uv} d(h_\theta(x_u), h_\theta(x_v)) + \sum_{n=1}^{L} c(g_\theta(x_n), y_n) \quad (13)$$

where $u, v$ are the seed node and its neighbour node and $w_{uv}$ is the edge weight between two nodes. $e_{ll}, e_{lu}$ and $e_{uu}$ refer to the graph edges in labelled-labelled, labelled-unlabeled and unlabeled-unlabeled. Where $h_\theta(x_u)$ refers to the hidden representation generated by the neural network with the input node as $x_u$, $d(h_\theta(x_u), h_\theta(x_v))$ is the distance between the seed node and the neighbour node's representation, $g_\theta(x_n)$ refers to the label prediction of the input $x_n$ by the network $g_\theta$ with weights as $\theta$, $c(\hat{y_n}, y_n)$ is the loss function of the neural network with the predicted label $\hat{y_n}$ and the target label $y_n$, and $\sum_{n=1}^{L} c(g_\theta(x_n), y_n)$ is the sum of network loss across the input data.

### 5) Mathematical Description of Our Method

Given a set of MRI representations as $V = \{v_1, v_2, ..., v_L\}$ and $L$ refers to the MRI image embedding vector size generated using either t-SNE, VAE, or AAE. The similarity between the nodes is measured using the cosine similarity measure in (14) to construct the similarity graph.

$$\text{Similarity} (v_i^l, v_j^l) = \frac{v_i^l \cdot v_j^l}{||v_i^l|| \times ||v_j^l||} \quad (14)$$

where vectors $v_i^l$ and $v_j^l$ are two different MRI representation vectors (i.e., nodes in a similarity graph) under the same label $l$, and the similarity ranges between -1 and 1. The constructed graph refers to $G = (V, E, W)$, and $E$ is the edges among nodes, and $W$ is the cosine similarity. Instead of using node-level information to construct the inference model, we apply a mapping function (i.e., embedding methods) to transform node-level information back to original MRIs, as shown in (15).

$$m(v_i, G, X) = x_i, \ x_i \in X, v_i \in G \quad (15)$$

where $x_i$ refers to the $i$th MRI image in the dataset $X = \{x_1, x_2 \dots x_L\}$. Except for the seed node $v_i$, the neighbours $u_i$ are also converted before feeding into the inference model. Then we apply the CNN model $f_\theta$ to minimize the distance between labels of seed MRI images and their neighbour as shown by (16) (i.e., unsupervised loss function of the CNN).

$$Loss_I = \underset{\theta}{argmin} \sum_{i=1}^{L} \sum_{j=1}^{N} w_{uv} || f_\theta(m(v_i, G, X)) - f_\theta(m(u_j, G, X)) ||^2 \qquad (16)$$

where $v_i, u_i \in G, u_i \in N(v)$, $C$ is the model weights, and $L$ is the node size, and $N$ is the max neighbours. $w_{uv}$ is the cosine similarity between nodes $v_i$ and $u_i$. $f_\theta(m(v_i, G, X))$ and $f_\theta(m(u_i, G, X))$ refers to the label prediction based on the MRI images of nodes $v_i$ and $u_i$. Equation (16) represents the CNN model's supervised part of predicting the seed MRI image labels as close as their neighbours. The neighbours must have a high similarity to the seed image to achieve fast convergence. The supervised part of the CNN loss function is shown in (17)

$$Loss_{II} = \underset{\theta}{argmin} \; \mathbb{E}[\mathcal{L}(f_\theta(m(v_i, G, X)), y_i)] \qquad (17)$$

where $\mathcal{L}(\cdot)$ is the CNN supervised loss function (sparse categorical cross-entropy) and $y_i$ is the actual MRI image labels. The sum of (16) and (17) is the total loss to be minimized. The final loss is shown as (18), and the weights of the CNN model can be found using the backpropagation algorithm.

$$\underset{\theta}{argmin} \{ \sum_{i=1}^{L} \sum_{j=1}^{N} w_{uv} || f_\theta(m(v_i, G, X)) - f_\theta(m(u_j, G, X)) ||^2 + \mathbb{E}[\mathcal{L}(f_\theta(m(v_i, G, X)), y_i)] \} \qquad (18)$$

Equation (18) first part "$\sum_{i=1}^{L} \sum_{j=1}^{N} w_{uv} || f_\theta(m(v_i, G, X)) - f_\theta(m(u_j, G, X)) ||^2$" is the unsupervised embedding loss function at CNN's target hidden layer. The second part "$\mathbb{E}[\mathcal{L}(f_\theta(m(v_i, G, X)), y_i)]$" is the categorical cross-entropy for supervised classification.

## IV. EXPERIMENTAL SETUP

In our study, we raise the following research question: "how can we instruct a convolutional neural network during training to maximize both structural similarity learning from the training data and the variance of the data that the CNN model can explain?". We further develop the following research hypothesis: if we can learn a structural similarity graph of the training data and then (1) use the structural similarity graph to present the input training samples with their k-nearest neighbours from the similarity graph to the CNN model during training, and (2) add the objective function of the similarity graph as an unsupervised loss function to the supervised loss function of the CNN model and minimize both losses together while training. As a result, we would be able to have a more accurate prediction of the testing data.

We further develop two experiments to test the research hypothesis, and both are implemented using TensorFlow 2 on the Google Colab cloud computing platform. The first experiment's objective is to compare the base CNN model's performance (without the similarity graph) to the graph regularized model when the dataset is significantly imbalanced. The first experiment also examines the effect of the similarity graph generated using various combinations of data representation and clustering algorithms on the model performance. The second experiment's objective is to further improve the first experiment's performance by utilizing a balanced dataset to examine the similarity graph regularization effect in this case.

## V. RESULTS AND DISCUSSION

Fig. 5 shows an illustration of our method to predict AD classes. The original dataset is separated into different categories based on labels. The MRI dataset has four labels include Non-Demented (ND), Very Mild Demented (VMD), Mild Demented (MD) and Moderated Demented (MDTD). The data pipeline starts with pre-processing, dimensionality reduction and similarity graph generation.

The pipeline continues with a full shuffle and splitting of MRI images into train and test (70% training and 30% testing). Eventually, the graph data is fed into the graph-regularized CNN model for the AD early-stage prediction.

The original MRIs (for a single image class label) are compressed into low-dimensional vectors using the t-SNE, VAE or AAE algorithm. The MRI representations are clustered using the K-Means or hierarchical clustering algorithm to reduce each cluster's data variance. The similarity graph is constructed for each image class based on the cosine similarity with a threshold. The threshold controls the edge strength among the nodes. Two nodes ($V$) in the similarity graph ($G$) will be connected by an undirected edge ($E$) if the cosine similarity between the two nodes is higher than a specific threshold. Each similarity graph has a mean node (cluster centroid), and all mean nodes are connected, linking all cluster graphs together.

The similarity graph describes the relationships among nodes by the edge weight and can control the training samples' assignment to CNN during training. Fig. 6 shows more details of our method for AD diagnosis using a similarity graph-regularized CNN model.

Fig. 6 shows an illustration of a CNN model training using a similarity graph to predict the four AD MRI image labels. The training input images are randomly sampled from the similarity graph with their k-nearest neighbours and fed to the CNN model. The 128-vector embeddings of the seed images and their k-nearest neighbours collected from the similarity graph are presented to the CNN model. The CNN model learns how to keep the distance between the seed embeddings and their k-nearest neighbour. Both inputs (an image and image's embedding) are considered simultaneously to make a final prediction, as shown in (18). The loss function is used to update the CNN model weights $\theta$ using the backpropagation algorithm. The similarity between the seed and its neighbours is calculated using the cosine similarity. The advantage of the cosine similarity is compressing the measurement into the range between -1 and 1.

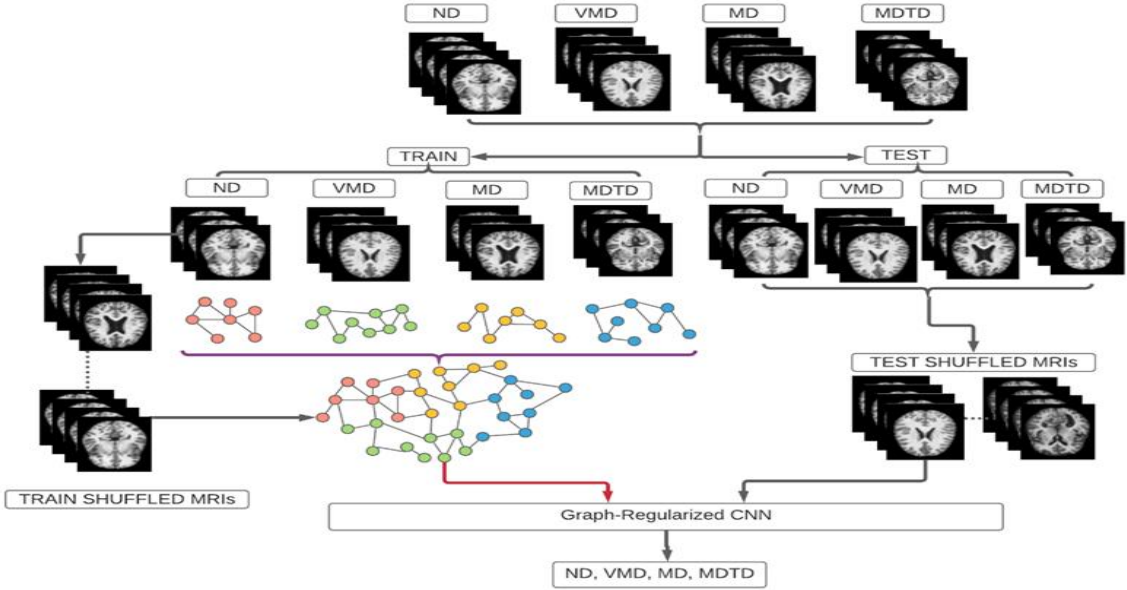

Fig. 5. An illustration of our method to predict AD at different stages.

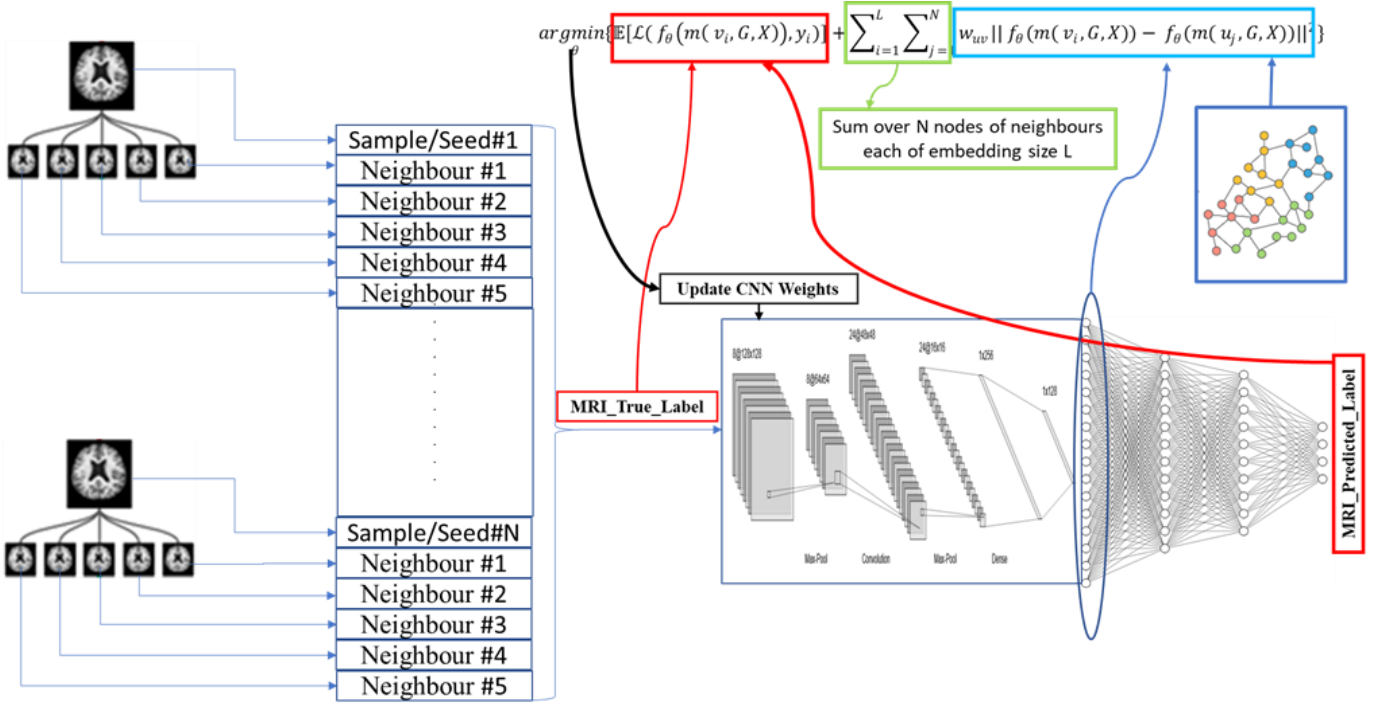

Fig. 6. CNN Graph-Regularized Model

The similarity threshold standardizes the similarity measurement in the similarity graph generation. However, the similarity measure calculates the cosine of the angle between two MRI representation vectors instead of the Euclidean distance between two vectors. Therefore, this calculation might not entirely reflect how similar two images are, and the similarity threshold has to be high enough to maintain a proper measurement of similarity between nodes. In our method, we choose a threshold of 0.75 to comply with the computation resources we have. The MRI image embedding vector representation is a compressed form of the original MRI images for a low computational cost. The data compression involves information loss and may fail to accurately reflect the original MRI data variance under each label. Therefore, a forced increase in the similarity threshold might cause a significant loss of seed nodes without neighbours.

We experiment with the K-Means and Hierarchical (HC) clustering algorithms to decrease each MRI category's variance to improve each cluster's similarity under each label. More clusters indeed improve the similarity between a seed MRI and its neighbours, but the number of seed MRIs decreases under each cluster. An improper similarity threshold might result in certain seed MRIs without neighbours in the same cluster under the same label. Therefore, seed nodes without neighbours are filtered out in the graph building process even though they are meaningful in the original MRI dataset. The similarity graph's target is to have high similarity on the node edges and maintain sufficient seed nodes and seed neighbours. The Seed Node Neighbour Count (SNNC) strongly relates to the Seed Node Count (SNC) of an image representation generated by AAE, VAE, and t-SNE models. The seed node of VAE decreases significantly when the similarity threshold increases up to 0.7. The VAE algorithm may reduce the MRI dimension, but compressed MRIs' variance is also changed under each label. In contrast, both seed nodes of AAE and t-SNE decrease by a small amount when maintaining a high similarity threshold at 0.95.

The t-SNE seed nodes have more neighbours across all seed nodes than AAE seed nodes have under the NonDemented label. The t-SNE algorithm representation has a lower embedding dimension than AAE. Thus, the t-SNE seed nodes have a smaller variance than the original MRIs, and all seed nodes are similar to each other. Therefore, MRI embedding representation using AAE maintains both dimension reduction and original MRI variance better than t-SNE and VAE representations.

Clustering is a useful step in decreasing the MRI embedding vectors' variance to create a better similarity graph and represent the MRI embeddings under each label and combined labels. However, it is challenging to judge the similarity threshold to generate the similarity graph representing the images' embeddings. Experiment I is conducted to explore the effect of the different settings used to generate the similarity graph without balancing the data, and experiment II is conducted with the data balancing and transfer learning techniques to boost the model performance.

### A. Experimental Results

#### 1) Experiment I

In this experiment, a VGG19 model is used as a base model to fit the imbalanced MRI image dataset, and the quality metrics such as Accuracy (ACC), Area Under the Receiver Operating Characteristics Curve (AUC) and F1score are collected to compare the model performance under each setting. The settings include various combinations of image representation and clustering algorithms to generate the similarity graph. The training parameters are consistent across all models, such as learning rates, optimizers, and batch size. The only difference is that the base CNN model uses MRI data, but the CNN similarity graph-based model uses MRI graph data (raw images and the raw images' embedding vectors).

Comparing different analytical processing pipelines is shown in Table II, and the best performance is achieved by combining the CNN model with a similarity graph, AAE and K-Means algorithm.

Experiment I can not directly compare graph accuracy among different pipeline settings, but the similarity graph quality can be reflected through the CNN model's performance. The CNN similarity graph-based model sees the seed MRIs and utilizes the seed neighbours' similarity to predict the imbalanced MRI image labels. The higher weights on the graph's edges could push the model prediction on the seed node in the right direction. Therefore, the graph generated using AAE with the K-Means algorithm can represent the MRI image embeddings with a high similarity threshold without a huge seed node loss.

The combination of the CNN similarity graph-based model, AAE and K-Means algorithm achieves the best performance at 71% at ACC, 86% at AUC and F1Score at 71%. The CNN similarity graph-based model (with AAE and K-Means algorithm) outperforms the base CNN model by 15% on ACC, 8% on AUC and 19% on F1Score when classifying the same MRI dataset. As shown in Table II, the model performance has a primary limitation due to the significantly imbalanced dataset. However, this reflects a significant advantage of the graph-regularized model in the imbalanced training dataset compared to the regular CNN model (base model). Therefore, when the dataset is balanced, the CNN similarity graph-based model's performance can be increased.

#### 2) Experiment II

In this experiment, we employ balancing techniques (data augmentation and bootstrapping) to balance the MRI image dataset and use it to fine-tune several pre-trained models, including VGG19, DenseNet121 (DN121), and Xception pre-trained models. The experimental results are shown in Table III. The DenseNet121 model with the similarity graph achieves the highest accuracy (98.6%) on the test dataset (70% train and 30% test), followed by Xception with the similarity graph (98.3%) and VGG19 with the similarity graph (98.2%).

The confusion matrix is shown in Fig. 7 for all models. The correct prediction of each label is shown on the diagonal of each matrix. For the ND label, all models most likely misclassify it as VMD label, which is the same for the MD label misclassification. In contrast, all models are more likely to misclassify the VMD label as the ND label. However, the MDTD label's prediction has the highest accuracy, which may be caused by balancing techniques. The CNN models can identify the MDTD label from others and only misclassify a few MD label images. Overall, the confusion matrix reflects that the prediction challenge is to generate clear boundaries between labels, especially ND vs. VMD, MD vs. VMD, and MDTD vs. (ND, VMD and MD).

Table IV shows each model performance (with and without the similarity graph) on each label prediction as a binary classifier. The DN121 similarity graph-based model achieves 98-99% overall ACC on each label classification. However, for each similarity graph-based model, the lowest classification ACC happens on the VMD labels because of the high overlap of features between the VMD, MD, and MDTD images. The primary limitation of model performance depends on how accurately the VMD label is classified.

The lowest recall rate happens on ND labels with the VGG19 similarity graph-based model (89.4%), VMD labels with DN121 similarity graph-based model (97.9%), and Xception similarity graph-based model (96.1%). Even though the DN121 similarity graph-based model achieves an overall ACC of 98% across all labels, this model can only achieve 97.9% accuracy when seeing MRIs with VMD labels, so the model performance is decreased by 1% in the binary classification task.

The lowest true negative rate is on VMD labels, with the VGG19 similarity graph-based model is 96%, while the DN121 similarity graph-based model predicts correctly almost 99% on VMD images. Therefore, both the recall and true negative rates reflect challenges to predict 'VD' labels when the models see MRIs with VD labels. However, the pre-trained fine-tuned models always have the best performance when seeing MRIs with MDTD labels.

The similarity graph-based models have 99% accuracy in predicting MRIs with MDTD labels, and this shows that bootstrapping with image augmentation can boost model performance. In contrast, MRIs with VMD labels are the primary limit in model performance. The DN121 similarity graph-based model can achieve 98% accuracy, 99.8% AUC and 97.2% F1Score. When this model works as a binary classifier, each label's prediction accuracy is 98.1% on ND, 97.9% on VMD, 98.2% on MD and 99.9% on MDTD. The results show that there is a significant difficulty in the VMD label prediction.

### B. Comparison to other related works

In this section, we compare our method against recent related works for AD diagnosis. A certain amount of research focuses on the binary classification of NC vs. AD. These researches contribute to the state-of-the-art methodology such as Siamese CNN [3], Multi-Channel CNN [4], and Inception V4 [25]. However, they ignore the conversion stages between

AD and NC, such as MCI, and its significance to diagnose this early-stage of AD and help patients prevent the symptom in the first place.

Therefore, this section compares our method against several recent studies in the early-stage prediction of AD. These studies are deep-learning based models and focus on the most challenging task of AD predictions such as AD vs. MCI, NC vs. MCI, or AD vs. NC vs. MCI. Our method deals with an even more challenging task since MRIs categorize MCI into VMD, MD, and end up with MDTD as the final stage before AD. Table V shows that our method outperforms other methods' performance.

As a binary classifier, the U-net-like CNN [11] performs better (97% ACC) when classifying NC and MCI. Our method achieves 95.9% accuracy when classifying MRIs with VMD labels. The ensemble learning method [10] may boost the performance, but it involves massive computation for fine-tuning the pre-training models. In contrast, our method leverages the benefits of transfer learning of state-of-the-art architecture. Our method utilizes the similarity graph as unsupervised learning to capture the similarity of the same class's images while maximizing the variance between images of different classes.

Fig.7. Confusion Matrix of Experiment II

TABLE II
EXPERIMENT I RESULTS

| Model Type | SNNC | ND vs. VMD vs. MD vs. MDTD | | |
|---|---|---|---|---|
| | | Acc(test) | AUC(test) | F1Score(test) |
| CNN base model | 0 | 0.56 | 0.78 | 0.53 |
| CNN + similarity graph + VAE | 5 | 0.57 | 0.78 | 0.48 |
| CNN + similarity graph + VAE + HC | 5 | 0.57 | 0.79 | 0.50 |
| CNN + similarity graph + VAE + K-Means | 5 | 0.57 | 0.82 | 0.54 |
| CNN + similarity graph + t-SNE | 5 | 0.57 | 0.83 | 0.53 |
| CNN + similarity graph + t-SNE + HC | 5 | 0.58 | 0.84 | 0.58 |
| CNN + similarity graph + t-SNE + K-Means | 5 | 0.56 | 0.83 | 0.52 |
| CNN + similarity graph + AAE | 5 | 0.60 | 0.84 | 0.58 |
| *CNN + similarity graph + AAE + K-Means* | *5* | *0.71* | *0.86* | *0.71* |
| CNN + similarity graph + AAE + HC | 5 | 0.58 | 0.83 | 0.46 |

TABLE III
EXPERIMENT II RESULTS

| Model | ND vs. VMD vs. MD vs. MDTD | | | | | |
|---|---|---|---|---|---|---|
| | SNNC | ACC(train) | AUC(train) | F1Score(train) | ACC(test) | AUC(test) | F1Score(test) |
| **VGG19** | 0 | 0.923 | 0.993 | 0.922 | 0.912 | 0.988 | 0.912 |
| **VGG19 + similarity graph** | 5 | **0.998** | 0.999 | 0.998 | **0.982** | 0.997 | 0.982 |
| **DN121** | 0 | 0.856 | 0.856 | 0.880 | 0.912 | 0.990 | 0.914 |
| **DN121 + similarity graph** | 5 | **0.996** | 0.999 | 0.996 | **0.986** | 0.998 | 0.987 |
| **Xception** | 0 | 0.855 | 0.978 | 0.852 | 0.901 | 0.987 | 0.898 |
| **Xception + similarity graph** | 5 | **0.994** | 0.999 | 0.994 | **0.983** | 0.997 | 0.982 |

TABLE IV
MODEL PERFORMANCE OF EXPERIMENT II

| Metrics | Labels | VGG19 | VGG19 + Similarity graph | DN121 | DN121 + Similarity graph | Xception | Xception + Similarity graph |
|---|---|---|---|---|---|---|---|
| **ACC (Accuracy)** | ND | 0.952 | 0.962 | 0.952 | 0.990 | 0.938 | 0.989 |
| | **VMD** | 0.924 | **0.953** | 0.917 | **0.983** | 0.915 | **0.984** |
| | MD | 0.959 | 0.980 | 0.956 | 0.992 | 0.927 | 0.992 |
| | **MDTD** | 0.988 | **0.999** | 0.999 | **0.999** | 0.952 | *0.999* |
| **ER (Error Rate)** | ND | 0.047 | 0.038 | 0.048 | 0.006 | 0.061 | 0.010 |
| | **VMD** | 0.075 | **0.047** | 0.082 | **0.013** | 0.085 | **0.016** |
| | MD | 0.041 | 0.020 | 0.043 | 0.008 | 0.048 | 0.008 |
| | **MDTD** | 0.012 | **0.001** | 0.002 | **0.000** | 0.003 | 0.000 |
| **Recall (TP rate)** | ND | 0.846 | **0.894** | 0.882 | 0.981 | 0.898 | 0.980 |
| | **VMD** | 0.916 | 0.931 | 0.918 | **0.979** | 0.845 | **0.961** |
| | MD | 0.938 | 0.964 | 0.990 | 0.982 | 0.861 | 0.961 |
| | **MDTD** | 0.951 | **1.000** | 0.999 | **0.999** | 0.997 | **0.990** |
| **Specialty (TN rate)** | ND | 0.990 | 0.985 | 0.976 | 0.994 | 0.953 | 0.993 |
| | **VMD** | 0.927 | **0.96** | 0.918 | 0.989 | 0.937 | **0.991** |
| | **MD** | 0.966 | 0.986 | 0.990 | **0.998** | 0.981 | 0.993 |
| | **MDTD** | 0.999 | **0.998** | 0.999 | **0.999** | 0.997 | 0.999 |
| **Fall Out (FP rate)** | ND | 0.009 | 0.015 | 0.023 | 0.005 | 0.046 | 0.007 |
| | **VMD** | 0.073 | **0.040** | 0.082 | 0.001 | 0.063 | **0.008** |
| | MD | 0.034 | 0.014 | 0.010 | **0.002** | 0.019 | 0.007 |
| | **MDTD** | 0.001 | **0.002** | 0.001 | **0.000** | 0.002 | **0.000** |
| **Miss Rate (FN rate)** | ND | 0.154 | 0.106 | 0.118 | 0.010 | 0.101 | 0.002 |
| | **VMD** | 0.084 | **0.069** | 0.082 | **0.020** | 0.154 | **0.039** |
| | MD | 0.062 | 0.004 | 0.150 | 0.025 | 0.138 | 0.011 |
| | **MDTD** | 0.049 | **0.000** | 0.000 | **0.000** | 0.003 | **0.000** |

TABLE V
COMPARISON WITH RECENT STUDIES IN AD EARLY-STAGE PREDICTION

| Model | Labels | Data Source | MRI Image | Accuracy. |
|---|---|---|---|---|
| CNN with a single MRI (2019) [26] | AD vs. s-MCI[6] | ADNI | 3D | 0.86 |
| DCNNs (Ensemble Learning) (2020) [12] | AD vs. MCI | ADNI | 3D | 0.94 |
| Multi-instance Learning with DCNN (2019) [11] | NC vs. MCI | ADNI | 2D | **0.97** |
| Multi-cluster Dense CNN (2017) [9] | NC vs. MCI | ADNI | 2D | 0.74 |
| Enhanced Inception Network (2019) [2] | AD vs. NC vs. MCI | ADNI | 2D | 0.85 |
| VGG (Transfer Learning) (2017) [27] | AD vs. NC vs. MCI | ADNI | 2D | 0.95 |
| Hybrid CNN (Ensemble Learning) (2019) [10] | AD vs. NC vs. MCI | OASIS | 3D | 0.95 |
| DCNNs with Multi-modality Images (2019) [28] | AD vs. NC vs. MCI | OASIS | 2D | 0.91 |
| *Our method: DN121+ similarity graph* | *ND vs. VMD vs. MD vs. MDTD* | **OASIS** | *2D* | *0.98* |

[6] s-MCI: stable- Mild Cognitive Impairment

## VI. CONCLUSION, LIMITATION, AND FUTURE WORKS

We propose a graph-based method to predict the early-stages of AD. Our method employs graph regularization to perform the supervised-learning for the AD classification. Our method is inspired by the NGL method and contributes several improvements in AD prediction researches as follows. The NGL method recommends using data embeddings or sparse feature representation to construct the similarity graphs [14]. However, our method constructs the similarity graph using dimension-reduction and clustering algorithms to generate MRIs' low-dimensional dense representations. Our method embeds the original MRI images into the similarity graph node to maintain detailed MRIs and the neighbouring relationships in the same graph. Our method gives a full workflow illustrating how the MRI similarity graph is constructed and used in the graph-regularized model and filling the gap in the application of NGL in the AD prediction research. Our method explores t-SNE, VAE, and AAE to reduce the MRI dimensions to construct the similarity graph. The graph generated using AAE and the K-Means algorithm reflects most of the original MRI distribution, and the similarity graph contributes to the model performance more than other discussed algorithms. Compared to previous methods in AD early-stage prediction, our method achieves excellent results using similarity graph-regularized learning. Also, we contribute to improving the NGL in supervised learning for the MRIs classification. However, there are some limitations to our method. Firstly, it is challenging to generate a standard threshold across the complete dataset to construct the similarity graph. Secondly, the graph is constructed as the undirected graph in the proposed method, but the directed graph could be more meaningful to represent AD's progression in different stages.